\documentclass[letterpaper,journal]{IEEEtran}
\usepackage{amsmath,amsfonts}
\usepackage{algorithmic}
\usepackage{algorithm}
\usepackage{array}
\usepackage[caption=false,font=normalsize,labelfont=sf,textfont=sf]{subfig}
\usepackage{textcomp}
\usepackage{stfloats}
\usepackage{url}
\usepackage{verbatim}
\usepackage{graphicx}
\usepackage{cite}
\usepackage[colorlinks=true,
            linkcolor=blue,
 citecolor=blue,
            urlcolor=blue]{hyperref}

\begin{document}

\title{SpectraTac: A Compact Camera-Free Optical Tactile Sensor with Distributed Color Sensing}

\author{     
 Hao Wu$^1$,~\IEEEmembership{Student Member,~IEEE,} Haotian Guo$^1$, Yu Feng$^1$,~\IEEEmembership{Member,~IEEE,} Yutong Wang$^1$,

Yanzhe Wang$^2$, Jianshu Zhou$^1$,~\IEEEmembership{Member,~IEEE}
\thanks{
Hao Wu and Haotian Guo contributed equally to this work.}
\thanks{This work was supported in part by the National University of
Singapore under the NUS Start-up Grant (FY2026). 1. Department of Mechanical Engineering, National University of Singapore, Singapore. 2. Department of Mechanical Engineering, Zhejiang University, Hangzhou, China. Corresponding author: Jianshu Zhou (e-mail: jianshuzhou@nus.edu.sg).    
}}

\markboth{}%
{}

\maketitle

\begin{abstract}

Tactile sensing is essential for physical interaction in robotics and human--machine systems. However, combining rich tactile information with compact hardware, low cost, and low computational overhead remains challenging. This work presents SpectraTac, a compact, camera-free optical tactile sensor that combines active red--green--blue (RGB) illumination with spatially distributed color sensing. Contact deforms a compliant transparent elastomer and modulates its internal light field, producing spatially differentiated changes in color and intensity. Three distributed color sensors capture these responses as low-dimensional spatio-spectral features, avoiding cameras, imaging optics, and high-dimensional image processing. The device measures 19.2~mm in diameter and 4~mm in height, with a material cost below USD~5. A data-driven decoding framework simultaneously estimates three-dimensional (3D) force and the contact region from the optical measurements. For 3D force prediction, the sensor achieved mean absolute errors (MAEs) of 0.161, 0.164, and 0.429~N along the $x$-, $y$-, and $z$-axes, respectively. The nine-region contact-classification accuracy was 99.9\%. We further evaluated real-time 3D force tracking and contact-region-based human--machine interaction through an interactive control task. These results indicate that distributed color-resolved optical sensing offers a compact, low-cost alternative to camera-based tactile sensing for robotics, wearable sensing, and interactive systems.

\end{abstract}


\section{Introduction}

\IEEEPARstart{T}{actile} perception provides direct information about contact, force, deformation, surface properties, and slip during physical interaction in robotics and human--machine systems. These quantities cannot always be inferred reliably from vision alone. Tactile sensing is therefore increasingly important for robotic grippers \cite{guo2025enabling, zhou2026everything, zhou2024dexterous} and dexterous hands \cite{wu2026dexlink,wu2026sylink}, where local contact feedback can improve grasp stability and adaptability during manipulation. A wide range of tactile sensors based on piezoresistive, capacitive, piezoelectric, magnetic~\cite{feng2026magcilia,guo2026magpillar}, and optical transduction have consequently been developed \cite{wang2025flexible, feng2026soft, feng2026bioinspired, wang2024gelsight}. Despite substantial progress, combining mechanical compliance, rich tactile information, compact packaging, low computational requirements, and straightforward integration remains challenging.

\begin{figure}[t]
    \centering
    \includegraphics[width=\columnwidth]{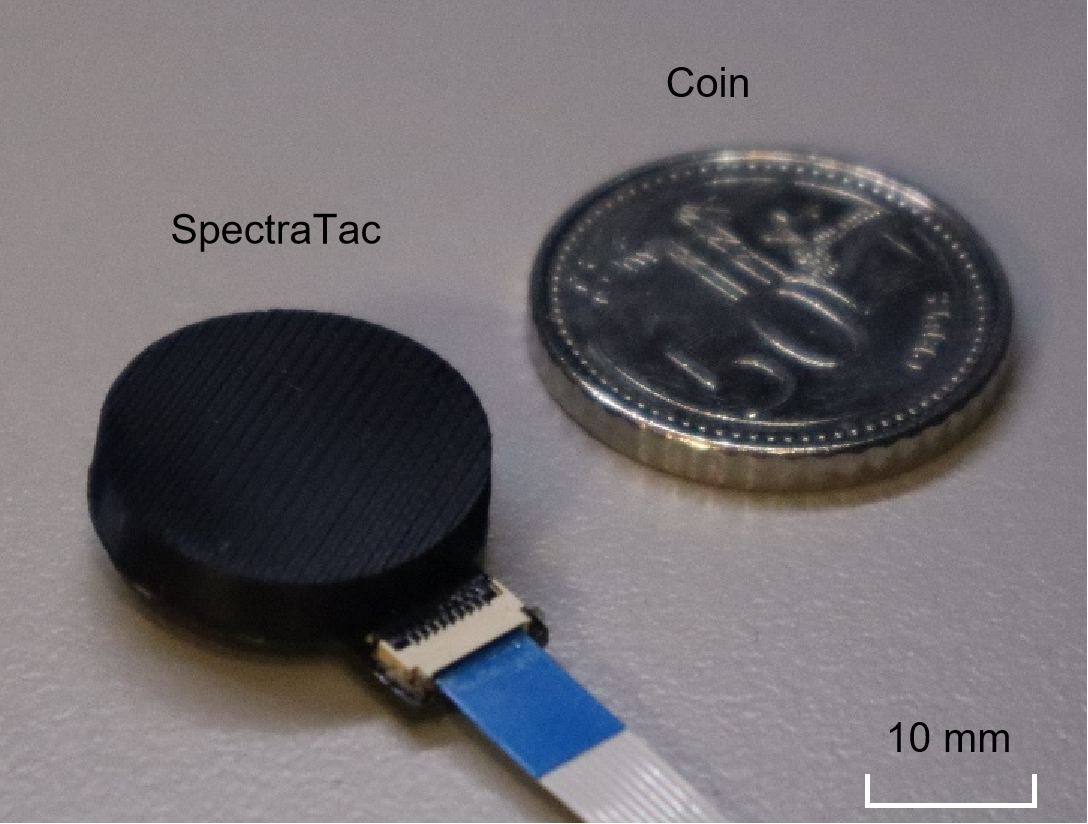}
    \caption{Prototype of the proposed compact camera-free optical tactile sensor.}
 \label{fig:prototype}
\end{figure}

Optical tactile sensing has attracted particular attention because deformation of a compliant medium can be converted into measurable optical signals. Camera-based sensors such as GelSight reconstruct contact geometry and force from images of a deformable elastomer \cite{yuan2017gelsight}, while biomimetic sensors such as TacTip infer contact from the displacement of internal markers \cite{ward2018tactip}. Subsequent designs, including GelSlim and DIGIT, have improved compactness and robustness for robotic manipulation \cite{donlon2018gelslim,lambeta2020digit}. Recent work has further explored multimodal sensing of soft robotic systems, leveraging visual observations and the deformation response of compliant structures to infer interaction forces and states \cite{wang2026relaxation}. Nevertheless, image-based tactile sensing requires a camera, imaging optics, sufficient optical path length, and dedicated image acquisition and processing. These requirements increase sensor size, data bandwidth, and computational demand.

Camera-free optoelectronic sensing provides an alternative approach by directly measuring deformation-induced changes in light propagation~\cite{wu2026lumitac}. Zhao \textit{et al.} demonstrated stretchable optical waveguides in which mechanical deformation modulates transmitted optical intensity, enabling curvature, elongation, and force sensing in a soft prosthetic hand \cite{zhao2016optoelectronically}. Piacenza \textit{et al.} embedded multiple light emitters and photodetectors within a transparent elastomeric layer. They used spatially overlapping light-transport signals to infer contact location and normal force over a multicurved robotic finger \cite{piacenza2020sensorized}. Optical encoding can further increase the information carried by a limited number of detector channels. Bai \textit{et al.} introduced stretchable distributed fiber-optic sensors incorporating chromatic patterns. The combined effects of absorption and frustrated total internal reflection distinguished the location, magnitude, and mode of deformation \cite{bai2020stretchable}. More recently, light direction and intensity have been jointly used to estimate 3D force and displacement in a compact tactile sensing architecture \cite{leslie2023tactile}. Optical-blocking structures have also enabled sensitive force measurements using simple light-emitting diode (LED)--photodiode arrangements for robotic grasping \cite{mo2025compact}. Together, these studies show that deformation-induced changes in optical transport can provide tactile information without full-field imaging.

The spectral dimension of light offers another way to increase tactile information density without substantially increasing sensor size. Unlike broadband intensity measurements, color-resolved optical responses provide multiple sensing channels for the same mechanical deformation. Yan \textit{et al.} developed soft, stretchable optical fibers containing gradient-colored segments. Their approach used wavelength-dependent attenuation and multiwavelength referencing for multipoint bending and pressure sensing \cite{yan2025soft}. Lee \textit{et al.} combined intensity-based optical waveguide sensing with colored regions. Changes in chromaticity distinguished local pressure at different positions while maintaining low computational complexity \cite{lee2025multimodal}. These studies show that color-resolved optical signals provide a richer compact representation than a single broadband intensity measurement, without requiring high-dimensional image acquisition. However, active multicolor illumination combined with spatially distributed color receivers remains underexplored in fingertip-scale tactile devices.

\begin{figure*}[t]
    \centering
    \includegraphics[width=1.0\textwidth]{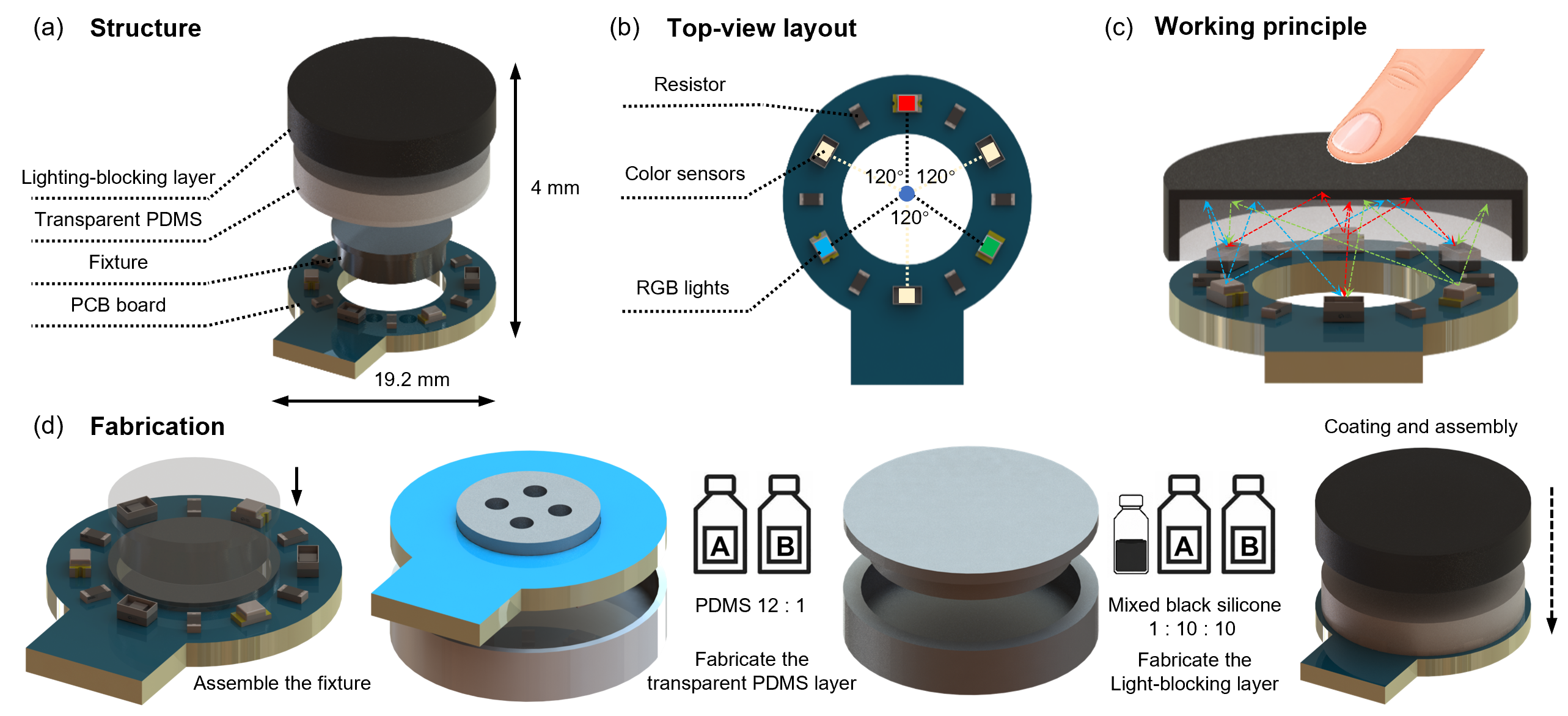}
 \caption{Structure, working principle, and fabrication of the proposed camera-free optical tactile sensor. (a) Exploded view of the structural components. (b) Top view of the printed circuit board (PCB), showing the alternating arrangement of RGBC sensors and RGB LEDs. Components of the same type are spaced at $120^{\circ}$ intervals, whereas adjacent emitters and sensors are separated by $60^{\circ}$. (c) Schematic of contact-induced modulation of internal light propagation. (d) Fabrication of the elastomeric layers and final assembly.}
    \label{fig:principle}
\end{figure*}

In this work, we present SpectraTac, a compact, camera-free optical tactile sensor (Fig.~\ref{fig:prototype}) that combines active RGB illumination with three spatially distributed color sensors. Contact-induced deformation modulates the internal light field within a compliant, transparent polydimethylsiloxane (PDMS) layer. This modulation generates spatially differentiated changes in color and intensity. The resulting low-dimensional spatio-spectral representation is decoded to estimate 3D force and the contact region without image acquisition or processing. The main contributions are threefold. First, we introduce a fingertip-scale, camera-free architecture that converts deformation-induced changes in internal light propagation into distributed color-resolved signals. Second, we develop an active spatio-spectral sensing scheme that combines RGB illumination with three red, green, blue, and clear (RGBC) sensors. The scheme encodes both force and contact-region information. Third, we integrate the compliant sensing medium, illumination, and color-detection electronics into a compact, readily manufacturable platform for robotics, wearable sensing, and human--machine interfaces.

\section{Sensor Design}

\subsection{Structural Design}
The proposed tactile sensor uses a compact, camera-free optical architecture comprising a transparent PDMS sensing layer, a light-blocking layer, three RGBC sensors, and three RGB illumination units, as shown in Fig.~\ref{fig:principle}(a). All optical components are integrated into a fingertip-scale cylindrical structure measuring 19.2~mm in diameter and 4~mm in height. This compact structure facilitates integration into robotic platforms and human--machine interfaces.

A key feature of the sensor is the alternating circumferential arrangement of three RGBC sensors and three RGB illumination units. As shown in Fig.~\ref{fig:principle}(b), components of the same type are symmetrically spaced at $120^{\circ}$ intervals. Adjacent alternating units are therefore separated by $60^{\circ}$. This distributed configuration senses contact-induced deformation from multiple directions and generates spatially differentiated optical responses. The transparent PDMS layer serves as both the compliant sensing medium and the optical propagation medium. RGB light propagates through the elastomer by internal reflection and scattering, with total internal reflection occurring when the critical-angle condition is satisfied. An opaque black elastomeric layer covers the sensing surface, suppresses ambient light and stray reflections, and deforms with the underlying PDMS.

\subsection{Working Principle}
Upon contact, elastomer deformation changes the local interface geometry, incidence angles, and optical path lengths, as illustrated in Fig.~\ref{fig:principle}(c). These changes redistribute the internal light field through variations in reflection, scattering, attenuation, and leakage. Owing to the distributed emitter--receiver arrangement, the three RGBC sensors produce distinct responses to different contact locations and loading conditions. Each sensor measures red, green, blue, and clear channels, and the response of the $i$th sensor is expressed as

\begin{equation}
\mathbf{s}_i =
\left[
R_i,
G_i,
B_i,
C_i
\right]^{\mathrm T},
\end{equation}

where $R_i$, $G_i$, $B_i$, and $C_i$ represent the measured red-, green-, blue-, and clear-channel intensities, respectively. Combining the outputs from the three spatially distributed sensors yields

\begin{equation}
\mathbf{x} =
\left[
\mathbf{s}_1^{\mathrm T},
\mathbf{s}_2^{\mathrm T},
\mathbf{s}_3^{\mathrm T}
\right]^{\mathrm T},
\end{equation}

The resulting vector forms a compact spatio-spectral representation of the contact-induced optical response.

The sensing mechanism combines spectral intensity information with spatially distributed responses. Applied forces deform the compliant optical medium and alter internal light propagation, producing characteristic intensity variations across the RGBC channels. The magnitude and pattern of these variations encode contact-induced deformation, while differences among the three sensors capture its spatial asymmetry. These complementary responses can therefore be decoded to estimate force magnitude, loading direction, and contact position. Unlike camera-based visuotactile sensors, the proposed sensor directly samples the deformation-modulated optical field using three miniature RGBC sensors. The alternating 60$^\circ$ emitter--receiver arrangement provides directional optical diversity without imaging optics, representing tactile information in a compact form factor.

\subsection{Fabrication}
The sensor was fabricated using commercially available, low-cost components and a simple molding process, as outlined in Fig.~\ref{fig:principle}(d). Three VEML3328 RGBC sensors were selected as the optical receivers because of their compact size and multichannel sensing capability. The transparent sensing layer was formed from PDMS by mixing the base and curing agent at a 12:1 ratio, providing optical transparency and mechanical compliance. The opaque shielding layer was prepared by mixing Ecoflex 00-30 at a 1:1 ratio and adding black silicone pigment. This layer suppressed ambient light and unwanted external reflections. After molding and curing, the elastomeric layers were assembled with the sensing and illumination components to form the complete device. The use of inexpensive commercial components and straightforward fabrication kept the total material cost below USD~5 per sensor, supporting scalable, low-cost fabrication.

\section{Tactile Signal Representation and Decoding}

\subsection{Ground-Truth Acquisition and Calibration}

\begin{figure*}[t]
    \centering
    \includegraphics[width=1.0\textwidth]{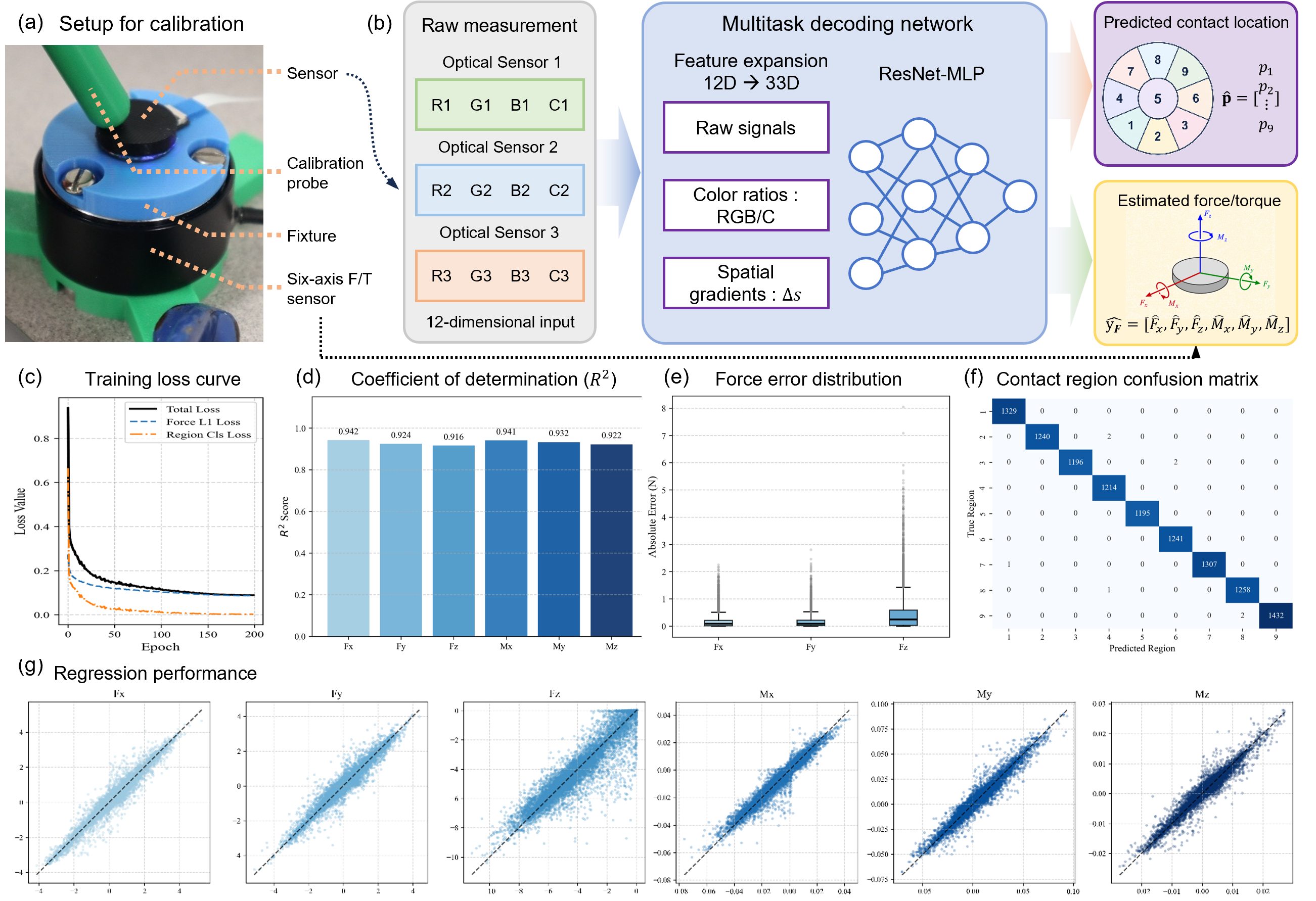}
 \caption{Data acquisition, decoding framework, and experimental evaluation. (a) Calibration setup using a reference six-axis F/T sensor. (b) Multitask learning pipeline that expands 12-dimensional raw optical signals into 33-dimensional features for F/T estimation and contact-region classification. (c) Training loss curves. (d) Coefficient of determination ($R^2$) for the six predicted mechanical components. (e) Absolute-error distributions for the 3D forces. (f) Confusion matrix for nine-region classification. (g) Predicted forces and torques versus their ground-truth values.}
    \label{fig:network}
\end{figure*}

During calibration, a six-axis force/torque (F/T) sensor was used as the reference for mapping optical measurements to tactile states. As shown in Fig.~\ref{fig:network}(a), the tactile sensor was mounted on the reference F/T sensor, and controlled contacts were applied to the sensing surface. The outputs of the three RGBC sensors were recorded synchronously with the corresponding six-axis F/T measurements. The reference mechanical state is represented as

\begin{equation}
\mathbf{y}_{F}=
\left[
F_x,
F_y,
F_z,
M_x,
M_y,
M_z
\right]^{\mathrm T}
\in \mathbb{R}^{6},
\end{equation}

where $F_x$, $F_y$, and $F_z$ denote the orthogonal force components, and $M_x$, $M_y$, and $M_z$ denote the corresponding torques.

In addition to the continuous six-axis F/T measurements, each contact sample was assigned a discrete label corresponding to one of nine representative regions on the sensing surface. These labels enabled simultaneous contact-region classification.

Each training sample comprises an optical measurement vector, a continuous F/T target, and a discrete contact-region label. The dataset is defined as

\begin{equation}
\mathcal{D}
=
\left\{
\left(
\mathbf{x}^{(n)},
\mathbf{y}_{F}^{(n)},
\mathbf{y}_{L}^{(n)}
\right)
\right\}_{n=1}^{N},
\end{equation}

where $N$ is the total number of samples, $\mathbf{x}^{(n)} \in \mathbb{R}^{12}$ is the optical measurement vector, $\mathbf{y}_F^{(n)} \in \mathbb{R}^6$ is the corresponding six-axis F/T vector, and $y_L^{(n)} \in \{-1, 1, \dots, 9\}$ is the contact-region label. Labels 1--9 correspond to the nine contact regions, whereas $-1$ indicates that no valid region annotation is available. Samples labeled $-1$ were retained for continuous F/T regression but excluded from the contact-region classification loss. In total, approximately 60,000 samples were collected over 50 minutes at 20~Hz. This protocol provided high acquisition throughput and a short calibration time.

\subsection{Physically Motivated Feature Representation and Normalization}
The sensor produces a 12-dimensional raw optical measurement vector $\mathbf{x}_{\mathrm{raw}} = [\mathbf{s}_1^T, \mathbf{s}_2^T, \mathbf{s}_3^T]^T \in \mathbb{R}^{12}$, and the decoding model estimates the six-axis mechanical state $\mathbf{y}_F \in \mathbb{R}^6$. 

To preserve the physical meaning of the measurements, feature expansion is applied directly to the raw optical intensities before statistical normalization. The 12-dimensional vector $\mathbf{x}_{\mathrm{raw}}$ is expanded into a 33-dimensional feature representation $\mathbf{x}_{\mathrm{enh}} \in \mathbb{R}^{33}$ comprising two physically motivated feature groups:

\begin{enumerate}
    \item \textbf{Clear-normalized chromaticity ratios}: To decouple spectral shifts from broadband intensity attenuation, the color ratios for each sensor $i \in \{1, 2, 3\}$ are computed as
    \begin{equation}
        \mathbf{c}_i = \left[ \frac{R_i}{C_i+\epsilon}, \frac{G_i}{C_i+\epsilon}, \frac{B_i}{C_i+\epsilon} \right]^T,
    \end{equation}
    where $\epsilon = 10^{-5}$ is a small positive regularization constant used to prevent division by zero under low-light conditions.
    \item \textbf{Pairwise sensor differences}: To represent the optical-flux asymmetry induced by tangential shear forces and multiaxis moments, pairwise differences between the distributed sensors are defined as
    \begin{equation}
        \Delta\mathbf{s}_{ij} = \mathbf{s}_i - \mathbf{s}_j, \quad (i,j) \in \{(1,2), (2,3), (3,1)\}.
    \end{equation}
\end{enumerate}

The complete 33-dimensional feature vector is formed by concatenating these feature groups:
\begin{equation}
    \mathbf{x}_{\mathrm{enh}} = [\mathbf{x}_{\mathrm{raw}}^T, \mathbf{c}_1^T, \mathbf{c}_2^T, \mathbf{c}_3^T, \Delta\mathbf{s}_{12}^T, \Delta\mathbf{s}_{23}^T, \Delta\mathbf{s}_{31}^T]^T.
\end{equation}
The resulting vector is standardized by z-score normalization across the training samples, yielding the network input $\tilde{\mathbf{x}} = (\mathbf{x}_{\mathrm{enh}} - \boldsymbol{\mu}_{\mathrm{enh}}) \oslash \boldsymbol{\sigma}_{\mathrm{enh}}$.

For the ground-truth mechanical targets $\mathbf{y}_F$, conventional mean-centering would map the resting-state reference to a nonzero normalized value. We therefore apply a zero-anchored scaling transformation:
\begin{equation}
    \tilde{y}_{F,j} = \frac{y_{F,j}}{\sigma_{F,j}}, \quad j=1,\dots,6,
\end{equation}
where $\sigma_{F,j}$ denotes the standard deviation of the $j$th F/T component, computed exclusively from the training set. Because no mean subtraction is applied, the zero-force and zero-torque reference remains at zero in the normalized target space.

\subsection{Shared-Backbone Multitask Learning and Masked Optimization}
As shown in Fig.~\ref{fig:network}(b), a shared-backbone residual multilayer perceptron (ResMLP) processes the 33-dimensional feature vector. The shared latent representation branches into a six-axis F/T regression head and a nine-class contact-region classification head. The classification head converts the logits into probabilities using a softmax function and selects the predicted contact region using the $\operatorname{argmax}$ operator.

To handle partially annotated data, the region label set includes $-1$ for samples without reliable contact-region annotations. We define the masked multitask objective as $\mathcal{L}_{total}=\mathcal{L}_{F}+\mathcal{L}_{L}$ and optimize it using a One-Cycle learning-rate schedule. The corresponding training-loss curves are shown in Fig.~\ref{fig:network}(c). The F/T regression head minimizes a component-weighted L1 loss, which assigns larger relative weights to the in-plane force components and smaller weights to the torque components after target scaling:
\begin{equation}
    \mathcal{L}_{F}=\frac{1}{6N}\sum_{i=1}^{N}\sum_{j=1}^{6}w_{j}|\tilde{y}_{F,i,j}-\hat{y}_{F,i,j}|,
\end{equation}
where $\mathbf{w} = [1, 1, 0.5, 0.1, 0.1, 0.2]^T$ corresponds to $[F_x, F_y, F_z, M_x, M_y, M_z]^T$. These coefficients prioritize selected components after target scaling and do not represent unit conversions. The contact-region classification head uses a masked cross-entropy loss $\mathcal{L}_L$. Samples with $y_L = -1$ are excluded from the classification loss but retained for F/T regression.

\begin{figure*}[htbp]
    \centering
    \includegraphics[width=1.0\textwidth]{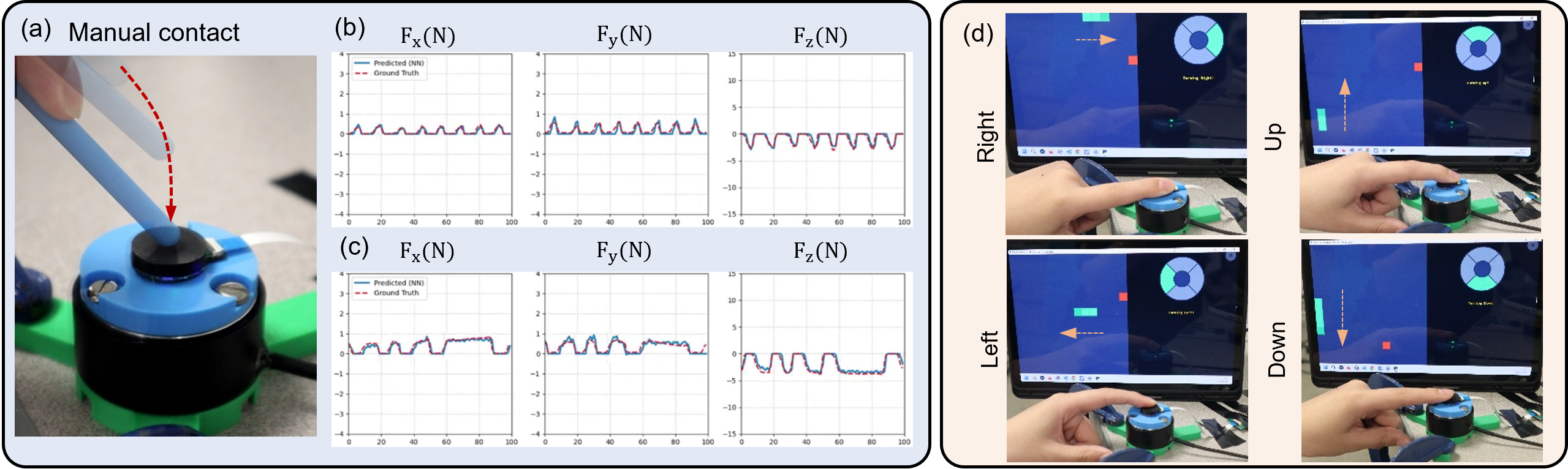}
    \caption{Real-time force estimation and human--machine interaction. (a) Dynamic manual contact applied to the sensor. (b) Tracking of 3D forces ($F_x$, $F_y$, and $F_z$) during repeated short contacts. (c) Tracking of 3D forces during sustained loads of different durations. In (b) and (c), solid blue lines show the predictions, and dashed red lines show the ground truth. (d) Real-time control of a Snake game. The nine contact regions are mapped to directional commands (up, down, left, and right), with the central region serving as the neutral state.}
    \label{fig:application}
\end{figure*}

\subsection{Decoding Performance}
The proposed representation and multitask learning (MTL) framework enabled simultaneous estimation of continuous forces and discrete contact regions. On the unseen test set, the model achieved $R^2 > 0.91$ for all six mechanical components, as summarized in Fig.~\ref{fig:network}(d). The component-weighted L1 loss, together with zero-anchored target scaling, reduced baseline drift. The MAEs were below 0.2~N for the in-plane shear forces ($F_x$ and $F_y$) and below 0.5~N for the normal force ($F_z$). The corresponding absolute-error distributions are shown in Fig.~\ref{fig:network}(e). For all three rotational axes, both the MAE and root-mean-square error (RMSE) were at most $0.004~\text{N}\cdot\text{m}$. Figure~\ref{fig:network}(g) summarizes the overall regression performance.

The masked classification head achieved a contact-region classification accuracy of 99.9\%. The remaining misclassifications occurred between adjacent regions, as shown by the confusion matrix in Fig.~\ref{fig:network}(f).

\section{Application and Demonstration}
We evaluated the sensor in two representative applications. The first assessed real-time 3D force estimation under continuously varying contact, whereas the second used contact-region classification for human--machine interaction through a tactile game interface. Both demonstrations used the same compact optical architecture to provide continuous force information and discrete contact-region information.

\subsection{Real-Time Three-Dimensional Force Estimation}
Real-time 3D force estimation ($\hat{\mathbf{F}} = [\hat{F}_x, \hat{F}_y, \hat{F}_z]^T$) was evaluated by applying continuously varying loads to the sensing surface (Fig.~\ref{fig:application}(a)). The trained model processed the distributed RGBC measurements, and its predictions were compared with simultaneously acquired reference F/T readings. As shown in Fig.~\ref{fig:application}(b) and (c), the predicted forces closely tracked the reference signals across the dynamic loading profiles. Figure~\ref{fig:application}(b) shows the response to repeated short contacts, whereas Fig.~\ref{fig:application}(c) shows stable tracking under sustained loads. By avoiding high-dimensional image processing, the proposed approach reduces computational overhead and supports continuous force feedback for robotic manipulation.

\subsection{Contact-Region-Based Human--Machine Interaction}
Beyond force estimation, the sensor served as a tactile interface in an interactive Snake game (Fig.~\ref{fig:application}(d)). The nine classification regions were mapped to directional commands. The central region served as a neutral state, while the peripheral regions were grouped into four directions (up, down, left, and right). To avoid accidental triggers, commands were executed only when the normal force exceeded $1.0\,\text{N}$.

During the interaction, a user pressed different peripheral areas of the sensing surface. The classification head continuously processed the distributed optical responses to predict the corresponding contact region. Predictions within one of the four designated zones were mapped to directional commands that steered the Snake in real time. Unlike conventional mechanical buttons or touchscreens, the proposed interface relied on deformation-induced optical responses within a compliant sensing surface, thereby preserving its mechanical flexibility.

This demonstration extends the sensing architecture beyond robotic force perception. The same low-dimensional optical signals supported continuous physical-state estimation and discrete interaction commands, suggesting potential applications in robotic control, handheld interfaces, and human--machine interaction systems.

\section{Conclusion}
This work presents SpectraTac, a compact, camera-free optical tactile sensor that encodes contact-induced deformation through distributed changes in color and intensity. Without image acquisition, its low-dimensional spatio-spectral representation supports accurate 3D force estimation and contact-region classification. The results establish distributed optoelectronic sensing as a compact alternative to camera-based tactile sensors for robotics and human--machine interaction. Future work will focus on improving spatial resolution, force decoupling, dynamic tactile perception, and long-term robustness. Direct optical readout may also enable the detection of dynamic events, such as slip and vibration, and support closed-loop tactile interaction.

\bibliographystyle{IEEEtran}
\bibliography{reference.bib}

\end{document}